# An Instrumented Wheel-On-Limb System of Planetary Rovers for Wheel-Terrain Interactions: System Conception and Preliminary Design

Lihang Feng

School of Instrument Science and Engineering, Southeast University, Nanjing 210096, China,

College of Control Science and Electrical Engineering, Nanjing Tech University, Nanjing 211816, China

fenglihang330@163.com

Xu Jiang

School of Instrument Science and Engineering, Southeast University, Nanjing 210096, China,

1429102892@qq.com

Aiguo Song*

School of Instrument Science and Engineering, Southeast University, Nanjing 210096, China,

a.g.song@seu.edu.cn

Understanding the wheel-terrain interaction is of great importance to improve the maneuverability and traversability of the rovers. A well-developed sensing device carried by the rover would greatly facilitate the complex risk-reducing operations on sandy terrains. In this paper, an instrumented wheel-on-limb (WOL) system of planetary rovers for wheel-terrain interaction characterization is presented. Assuming the function of a passive suspension of the wheel, the WOL system allows itself to follow the terrain contour, and keep the wheel remain lowered onto the ground during rover motion including climbing and descending, as well as deploy and place the wheel on the ground before a drive commanding. The system concept, functional requirements, and pre-design work, as well as the system integration, are presented.

## 1 INTRODUCTION

Wheeled mobile robots (WMR) have been deployed in Mars/Lunar exploration, military missions, and geological investigations, and they must handle rough and deformable terrains. The rovers have faced significant difficulty when traversing the Martian/Lunar surface because of the wheel slip and sinkage [1, 2]. The most notable example is the NASA Spirit which was immobilized by its trapped wheel in April 2009 [3]. Therefore, understanding the wheel-terrain interaction (WTI) is of great importance to improve the rover traversability.

The dynamics at the wheel–soil interface are major factors in the entrapment of WMR and the subsequent mission failure. To characterize the wheel-terrain interaction, many terramechanics based models have been developed to estimate these wheel-terrain interactions, including the Bekker model [4], the Wong-Reece model [5] and its extended variations developed by Shibly et al. [6], Iagnemma et al. [7], and Ding et at. [8]. However, the model-based estimation were relying on the assumption of ideal conditions with homogeneous soil and flat terrain, and many internal parameters such as the soil properties and shear deformation are required to quantify the wheel slip and sinkage. For the in-situ sensing of terrain characteristics, Iagnemma et al. [9] performed a multi-sensor measurement on a rover using information from the wheel motors and camera modules mounted outside the wheel. Ojeda et al. [10] investigated terrain characterization for a skid-steer robot using motor power consumption to estimate the wheel force and the encoder to compute the wheel rotational angle. The methods are generally indirect measurements via the proprioceptive sensors within the uncertain interferences.

From the viewpoint of planetary traverses, the concepts [11] to reduce the risk of immobilization failure heavily rely on human involvement and simulation of rover operations including building up a 3D environment of the current surroundings, identifying the hazards with manual recognition of the suspected subsurface hazards or high slip, and then planning and validating the paths. A well-developed sensing device carried by the rover would greatly facilitate the complex risk-reducing operations. The concept of instrumented wheel deployed in front of the robotic rover to provide representative terrain loading information is very promising, including the Wheeled Bevameter (WB) [12] and the Legged PathBeater (LPB) [13]. The WB is a sensing wheel that originates from the traditional bevameter in terrestrial trafficability test for off-road vehicles and became popular due to the ExoMars project [11]. The LPB is a sensor concept that comprises two arms with pyramidal penetrators at the tips mounted on top of the rover wheels, so it can use inertial sensors and strain gauges to estimate the bearing strength. However, these state-of-art developments are still far away from the real equipment or payloads carried by Mars/Lunar rovers although some are directly intended for the application of planetary exploration missions. Another issue is that many of these works are not publicly disclosed so that few documents of technique details cannot be acquired.

In this paper, the preliminary evaluation and design for an instrumented wheel-on-limb system installed on planetary rovers are presented. It aims for the potential in-situ sensing and detecting of wheel-terrain interactions before the traversal of the robotic vehicle. The paper is organized as follows. Section 2 describes the system concept of the wheel-on-limb mechanism. Section 3 elaborates the design, implementation, and mechatronics integration. Section 4 demonstrates the system with calibration and experimental tests.

## 2 SYSTEM CONCEPT

### 2.1 Desired Design Requirements

In future planetary missions, the need for safe, long, and sustained traverses on the surface is much greater than for the current rovers. The proper instrument that provides additional capabilities should be enabling planetary rovers to reliably and rapidly traverse long distances over unknown terrain. The state-of-art techniques involving the terrain traversability on planetary rovers are heavily relying on the vehicular camera. For example, the Zhurong rover [14] has carried the Navigation and Terrain Cameras (NaTeCams) for environmental imagery information aquisition in terrain identification and analysis. The camera-based terrain data helps to construct topography maps, extract parameters such as slope, undulation, and roughness, and investigate geological terrain structures. However, the non-geometrical accidents such as wheel slip and sinkage are still not well addressed, although the ground verification of several rover configuration designs has been implemented. Therefore, developing a direct wheel-terrain sensing device carried by the rover, rather than manually operation for terrain traversability analysis relied solely on the vehicular camera, would greatly facilitate the complex risk-reducing operations in future long and safe planetary traverses. The proposed system herein is intended for the validation of future long-safe planetary traverses and will be potentially used as a vehicular instrument targeted for the payload on the future planetary rovers. The desired design requirements for functional description are as follows.

1) The primary function of such an instrumented system is to directly measure the real-time dynamics at the wheel-terrain interface including the wheel forces, the contact angle, and wheel sinkage in field conditions.

2) An instrument payload carried on the rover should be easily deployed and placed to probe into the non-geometric hazard that was invisible by NaTeCams, so the forward sensing of terrain trafficability can be realized.

3) The rover's scientific exploration with low traversing speed (e.g. Zhurong rover is about 1.1cm/s) makes it possible to get the terrain information immediately ahead of the rover traversal, so the rover will have the potential ability for path re-planning and global hazard map updating beyond the immediate range of navigation.

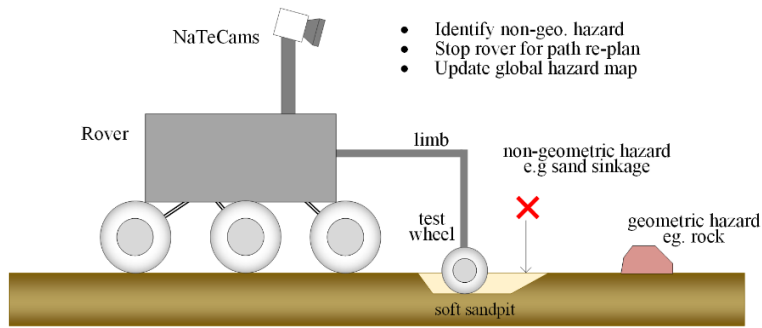

Figure 1: Concept of the Instrumented Wheel-On-Limb system for in-visible hazard detection on soft terrain sandy trap.

Mechanism Configuration: As inspired by the wheel-on-limb mechanism, the system design adopts a deployment and placement mechanism (limb) that is holding an instrumented wheel, located at front of the robotic vehicle (see Figure 1) to in-situ sense the terrain's physical properties, so it can provide the additional information (e.g. the non-geometric hazard such as sinkage) immediately in the vicinity of a front area ahead of the rover traversal. Thus, assuming the function of a passive suspension of the wheel, it should 1) allow itself to follow the terrain contour, and 2) keep the wheel remain lowered onto the ground during rover motion including climbing and descending, as well as 3) deploy and place the instrument wheel on the ground when it is required before a drive commanding. Table.1 shows the target specification regarding the functional requirement. An instrumented wheel is also required as a testing device on the terrain to enable the real-time quantities measurement. Generally, slip and sinkage are the main concerns that happen at the wheel-terrain interface, and fundamentally, they are the dynamic effects driven by the wheel. The dynamical forces generated on the terrain surface are the concentrated drawbar pull, the normal force, and the wheel-terrain reaction torque, respectively. Such an instrumented wheel can be a wheel force transducer/sensor (WFT) referred to as our previous work [15, 16]. Herein, this work is intended for the conventional evaluation and preliminary design of the limb mechanism to support the instrumented wheel.

Table 1: Functional Requirements and Design Specifications

| Mechanism | Function & Design specifications |
|---|---|
| Limb motion | 1) Deploy, place, and lift operation: ≥2 DOF motions.<br>2) Terrain-contour following: [-30°, 50°]degrees around Y-axis<br>3) Loading constraints: falling into the scale of WFT measurement<br>4) Dimension constraints: Limb length ≈ 1/2 rover length |
| Wheel motion | free-rolling as a passive suspension of the wheel:<br>1) free-rolling around Y-axis;<br>2) controlled rotation around Z-axis. |

**2.2 Conceptual Design and Simulation**

As we have assumed the function of a passive suspension of the wheel, the comprehensive concept of the design target in a representative simulation environment can be performed. To make the holding-wheel follow the terrain contour, the DPM limb can be intuitively designed as a parallelogram mechanism which results in a rotation movement around y-axis with respect to the rover frame. The simulation herein is to validate the conceptual design of the Wheel-On-Limb measurement system, especially for the limb.

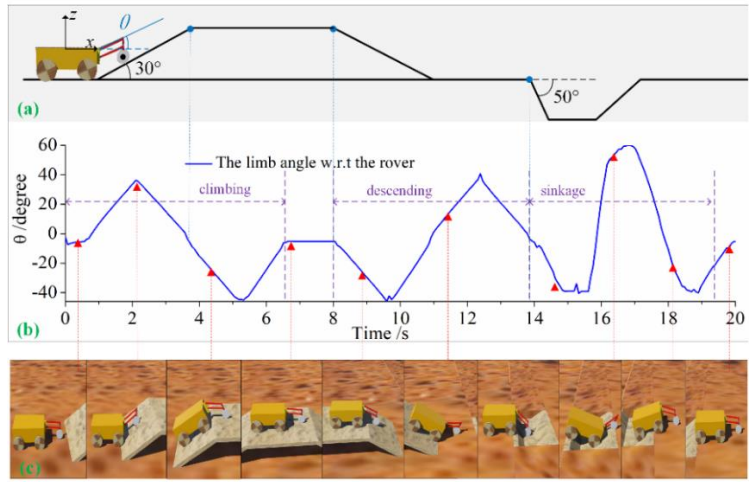

Figure 2: Simulation of the Wheel-On-Limb system: (a) simulated terrain contour; (b) the limb angle with respect to the rover; (c) screenshots of different rover poses in Webots simulator.

As shown in Figure 2(a), we simulate the Wheel-On-Limb system in a representative environment including the working conditions of climbing, descending, and sandy-trap sinkage via the open-source robot simulator Webots [17]. The most concerning parameter i.e. the limb angle $\theta_l$ of the rover (read from the Webots simulator) can be shown in Fig.2.b. It gives the intuitive variations of $\theta_l$ (-40°~60°) during the movement to guarantee the required simulation conditions e.g. climbing and descending angle of 30°, sandy-trap angle of 50°, and sinkage depth of 100 mm. Meanwhile, the limb parameters i.e. the lengths of two adjacent sides for the parallelogram mechanism ($l_1$ and $l_2$ in Figure 3) can be also determined in such a simulation environment.

## 3 DESIGN, IMPLEMENTATION, AND SIMULATION

This section presents the design and implementation of the proposed wheel-on-limb measurement system. According to the design specifications, the limb mechanism (see Figure 3) consists of a parallelogram structure to support the terrain-contour following and a four-bar linkage mechanism to provide the motion driving.

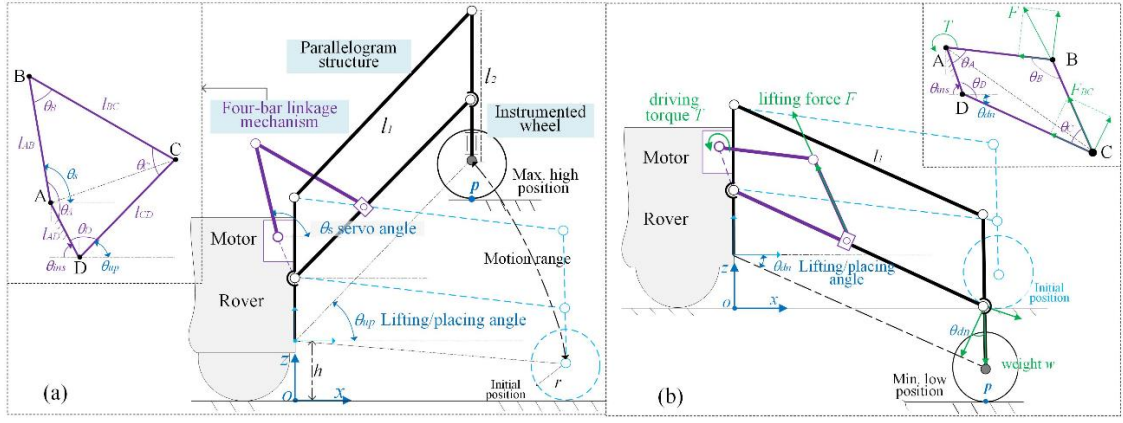

Figure 3: Limb mechanism with lifting state (a) and placing state (b)

### 3.1 Kinematics Analysis

As shown in Figure 3(a), the kinematic for the parallelogram structure presents the relationship between wheel position $\boldsymbol{p}$ and the lifting angle $\theta_l$, expressed as

$$\boldsymbol{p} = \begin{bmatrix} p_x \\ p_z \end{bmatrix} = \begin{bmatrix} l_1 \cos\theta_l \\ l_1 \sin\theta_l + h - r \end{bmatrix} \quad (1)$$

where $l_1$, $h$, and $r$ are all constant parameters as shown in Figure 3. The angle $\theta_l = \{\theta_{up}, \theta_{dn}\}$ is constrained by the motion range corresponding to the lift-folding state ($\theta_{up}$) and terrain-following state ($\theta_{dn}$). Thus, the limb has a workspace around y-axis within $[-40, 60]$ degrees, and the wheel rotates around z-axis within $[0, 90]$ degrees. The kinematic for the four-bar linkage mechanism shows the driving angle $\theta_s$ as a function of the lifting angle $\theta_l$ and an adjustable link $l_{CD}$. From the inset figure of Figure 3(a), it can be computed by

$$\theta_s = f_{kin}(\theta_l, l_{CD}) = \theta_A - \theta_{ins} = \angle BAC + \angle DAC - \theta_{ins} \quad (2)$$

where $\theta_{ins}$ is the installation angle of the bar link $l_{AD}$. By solving the triangle $\triangle BAC$ and $\triangle DAC$ within the cosine law, respectively, we have

$$\begin{cases} \angle BAC = \arccos\left(\frac{l_{AB}^2 + l_{AC}^2 - l_{BC}^2}{2 l_{AB} l_{AC}}\right) \\ \angle DAC = \arccos\left(\frac{l_{AD}^2 + l_{AC}^2 - l_{CD}^2}{2 l_{AD} l_{AC}}\right) \end{cases} \quad (3)$$

where $l_{AB}$, $l_{BC}$, and $l_{AD}$ are all constant variables, leaving the only unknown $l_{AC}$ that can be computed by

$$l_{AC} = \sqrt{l_{AD}^2 + l_{CD}^2 - 2 l_{AD} l_{CD} \cos\theta_D} \quad (4)$$

where $\cos\theta_D = \cos(\pi - \theta_i - \theta_{ins})$. Substituting Eq.(3) and (4) into Eq.(2), it can be used to find the optimal solution of adjustable link $l_{CD}$ (as a parameter) and the servo angle $\theta_s$ (as a kinematic variable) in the design phase. After the optimization, the final kinematics relationship can be acquired by substituting the inverse functions $\theta_i = f_{kin}^{-1}(\theta_s)$ into Eq.(1), it yields

$$\boldsymbol{p} = \begin{bmatrix} p_x \\ p_z \end{bmatrix} = \begin{bmatrix} l_1 \cos[f_{kin}^{-1}(\theta_s)] \\ l_1 \sin[f_{kin}^{-1}(\theta_s)] + h - r \end{bmatrix} \quad (5)$$

Therefore, it is easy to make sure the mechanism parameters and specifications meet the design requirements of kinematics in the design phase.

### 3.2 Force-Transmission Analysis

The force transmission is to analyze the driving torque $T$ as a function of the parameters $(\theta_i, l_{CD})$. It helps to determine the motor specification and link strength by changing the different parameters. Taking the Figure 3(b) as an example, force equilibrium equations can be expressed

$$\begin{cases} T_s = F_{BC}\sin(\pi - \theta_B) = -F_{BC}\sin\theta_B \\ l_{CD} \cdot F_{BC}\sin\theta_C = l_1 \cdot w\cos\theta_l \end{cases} \quad (6)$$

and it yields

$$T_s = -\frac{l_1 w \cos\theta_i \sin\theta_B}{l_{CD}\sin\theta_C} = f_2(\theta_l, l_{CD}) \quad (7)$$

where the internal angles in Eq.(7) can be given by

$$\begin{cases} \theta_B = \arccos\left(\frac{l_{AB}^2 + l_{BC}^2 - l_{AC}^2}{2l_{AB}l_{BC}}\right) \\ \theta_C = \arccos\left(\frac{l_{AC}^2 + l_{BC}^2 - l_{AB}^2}{2l_{AC}l_{BC}}\right) + \arccos\left(\frac{l_{AC}^2 + l_{CD}^2 - l_{AD}^2}{2l_{AC}l_{CD}}\right) \end{cases} \quad (8)$$

Eq.(8) can be used to find the maximum torque required for the servo motor in the design phase. Note that when the torque $T$ is applied in opposite direction, it provides the loading force of the wheel on the ground.

### 3.3 Simulation and Optimization

From the above-mentioned functions, the optimal solutions of the limb design can be computed and analyzed by the parameter response simulation. Note that there are many determined links owning to the robotic geometrical constraints. A final list of the parameters is shown in Table 2. The link $l_1$ in the parallelogram and the links $\{l_{AD}, l_{AB}, l_{BC}\}$ in the four-bar linkage mechanism are self-defined to meet the size of the test robotic vehicle in the subsequent section (Autolabor Pro1, Tsinghua University). Note that during such a simple optimization process, the primary servo motor is a pressing concern because it drives the limb to conduct the lifting and placing operation. A dedicated simulation is processed to determine the motor selection. As shown in Figure 4, the driving angle $\theta_s$ and torque $T$ are plotted as functions of the sensitive variables (i.e. $\theta_i$ and $l_{CD}$). The blue dot line in Figure 4 shows the moderate option for such a design. A final value of about 160mm for link $l_{CD}$ corresponding to a minimum value $\theta_s$ about 45.6 degrees and corresponding to a maximum torque $T_s$ about 2.8N.m is obtained. Note that the driving torque is computed within the gravity load $w$ as a standard reference in the force-transmission computation, so the computed result should be scaled to the specification of real conditions in the system integration.

Table 2: Parameters and optimal values in computation for the limb mechanism

| Parameters | $l_{AD}$ | $l_{AB}$ | $l_{BC}$ | $l_1$ | $l_{CD}$ | min{$\theta_s$} | max{$T_s$} | w |
|---|---|---|---|---|---|---|---|---|
| value | 102mm | 120mm | 130mm | 370mm | 160mm | 45.64° | 2.8N.m | 9.8N·m |

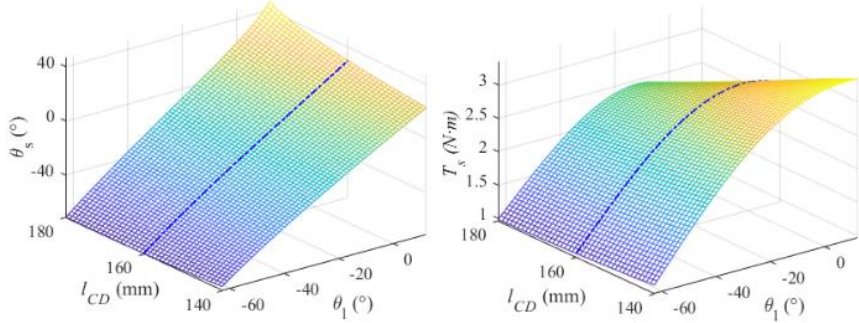

Figure 4: An example of the simulated optimization results for (a) the driving angle parameter and (b) driving torque. The blue dot line is an optimal result in the simulation

## 4 SYSTEM INTEGRATION AND VERIFICATION

### 4.1 System Integration

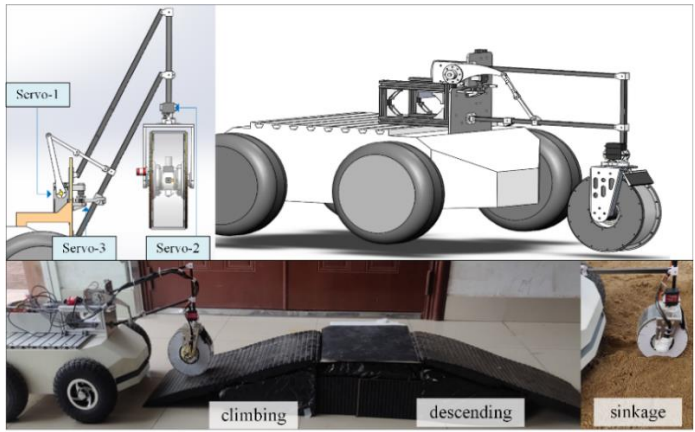

Figure 5: Mechatronics assembly and system validation of the proposed Wheel-On-Limb system

Figure 5 shows the overall Wheel-On-Limb system installed on a pre-assigned robotic vehicle. It can be seen that the limb part acts as a driving mechanism and the test wheel works as a sensing device. The limb mechanism is controlled by three servo motors. Servo-1 is a high-power servo motor for the limb lifting and placing which supports a steady-state torque of 500kg.cm at the rated speed of 0.12s/60°. Servo-2 has a steady-state torque of 20kg.cm at the rated speed of 0.12s/60° and it drives the test wheel to rotate around z-axis. Servo-3 is only 2kg.cm to work like a locker of the limb when it is not working. The proposed system is composed of a power supply module, sensor module, and control module. The power module receives the power from the robotic vehicle by wire connection and provides stable electric power for each module. The instrumented wheel enables the online wheel-terrain interaction measurements. The control module processes the information through a microcontroller unit obtained from the sensors and interact with an upper computer where the signal can be processed. It also accepts the commands from the upper computer and drives the motors to move.

### 4.2 Functional Validation

Functional validation of the proposed system is performed in a simple indoor and outdoor environment. As shown in Figure 5, the proposed Wheel-On-Limb mechanism works as a passive suspension of the wheel, so we can intuitively observe that the rover can drive the test wheel to follow the terrain contour with climbing, descending, and simulated sinkage trap. We also observe that the limb mechanism works well to support the test wheel with the lifting and placing operation controlled by the motors. Table3 shows the functional requirement validated by the real rover test. The comprehensive concept of the design target can be intuitively realized. An average motion accuracy can be observed and computed within several times real test. Results showed an accuracy of about [1.8mm, 3.7mm] for translation in x-y plane and of about [0.04rad, 0.08rad] for rotation in y-z plane. The relatively large translation error in the y-direction is owning to the "terrain contour-following", however, such a variation is acceptable compared to the limb movement between [-40°, 60°] when the wheel contacts with the terrain.

Table 3: Targeted functional validation

| | Design specifications | Target items |
|---|---|---|
| Limb motion | 1) Deploy, place, and lift operation: ≥2 DOF motions.<br>2) Terrain-contour follow: [-30°, 50°] around Y-axis<br>3) Loading const.: falling into the scale of WFT<br>4) Dimension const.: Limb length ≈ 1/2 rover length | 1) 2-DOF motion with three actuators<br>2) [-40°, 60°] around Y-axis<br>3) Yes, constrained by WFT design<br>4) Length: Limb=370mm , rover=726mm |
| Wheel motion | free-rolling as a passive suspension of the wheel:<br>1) free-rolling around Y-axis;<br>2) controlled rotation around Z-axis. | Instrumented wheel:<br>1) [0°, 360°] around Y-axis<br>2) [0°, 90°] around Y-axis with Servo-2 |

### 5 CONCLUSION AND DISCUSSION

In this paper, the preliminary evaluation and design for an instrumented wheel-on-limb system installed on the planetary rovers are presented. The system concept, functional requirements, pre-design work, robotic simulation, kinematics, and force analysis as well as the system integration are presented. Assuming the function of a passive suspension of the wheel, the proposed wheel-on-limb system can allow itself to follow the terrain contour, and 2) keep the wheel remain lowered onto the ground during rover motion including climbing and descending, as well as place the wheel on the ground when it is required before a drive commanding.

The current work is part of the preparatory project supported by the China Academy of Space Technology involving the next generation of planetary rovers for safe, long, and sustained traverses missions. The system is intended for the direct solution of characterizing wheel-terrain interactions prior to the traversal of the robotic vehicle, while the current work only shows the preliminary design and evaluation. One clear drawback is that we used a four-wheel robot rather than a rocker-bogie rover to test the proposed system, even though the accuracy seems satisfactory. Much follow-up progress should be done to improve the current system in future work.

**ACKNOWLEDGMENTS**

This work is supported by the grant of National Natural Science Foundation of China (62103184) and China Postdoctoral Science Foundation (2021M690630).